\documentclass[10pt,twocolumn,letterpaper,final]{article}
\usepackage{cvpr}              

\usepackage[dvipsnames]{xcolor}

\definecolor{cvprblue}{rgb}{0.21,0.49,0.74}
\usepackage[pagebackref,breaklinks,colorlinks,citecolor=cvprblue]{hyperref}

\def\paperID{401} 
\def\confName{CVPR}
\def\confYear{2024}

\usepackage{float}
\usepackage{graphicx}
\usepackage{verbatim}       
\usepackage[linesnumbered,ruled,vlined]{algorithm2e}
\usepackage{algorithmic}

\title{BEV-IO: Enhancing Bird's-Eye-View 3D Detection with Instance Occupancy}
\author{Zaibing Zhang$^{1}$ \quad Yuanhang Zhang$^{1}$ \quad  Lijun Wang$^{1}$ \quad Yifan Wang$^{1}$ \\ Huchuan Lu$^{1}$\\
$^{1}$ Dalian University of Technology, China \\
}
\begin{document}
\maketitle
\begin{abstract}
A popular approach for constructing a bird's-eye-view (BEV) representation in 3D detection is to lift 2D image features onto the viewing frustum space based on explicitly predicted depth distribution. However, depth distribution can only characterize the 3D geometry of visible object surfaces but fails to capture their internal space and overall geometric structure, leading to sparse and unsatisfactory 3D representations. To mitigate this issue, we present BEV-IO, a new 3D detection paradigm to enhance BEV representation with instance occupancy information. At the core of our method is the newly designed instance occupancy prediction~(IOP) module, which aims to infer point-level occupancy status for each instance in the frustum space. To ensure training efficiency while maintaining representational flexibility, the IOP module is trained by combining both explicit and implicit supervision. 
With the predicted occupancy, we further design a geometry-aware feature propagation mechanism~(GFP), which performs self-attention based on occupancy distribution along each ray in frustum and can efficiently aggregate instance geometry cues into image features. By integrating the IOP module with GFP mechanism, our BEV-IO detector is able to render highly informative 3D scene structures with more comprehensive BEV representations. Experimental results demonstrate that BEV-IO can outperform state-of-the-art methods while only adding a negligible increase in parameters ($0.2\%$) and computational overhead ($0.24\%$ in GFLOPs).
\end{abstract}    
\section{Introduction}
\label{sec:intro}

\begin{figure}[!ht]
    \begin{center}
    \includegraphics[width=1.1 \linewidth]{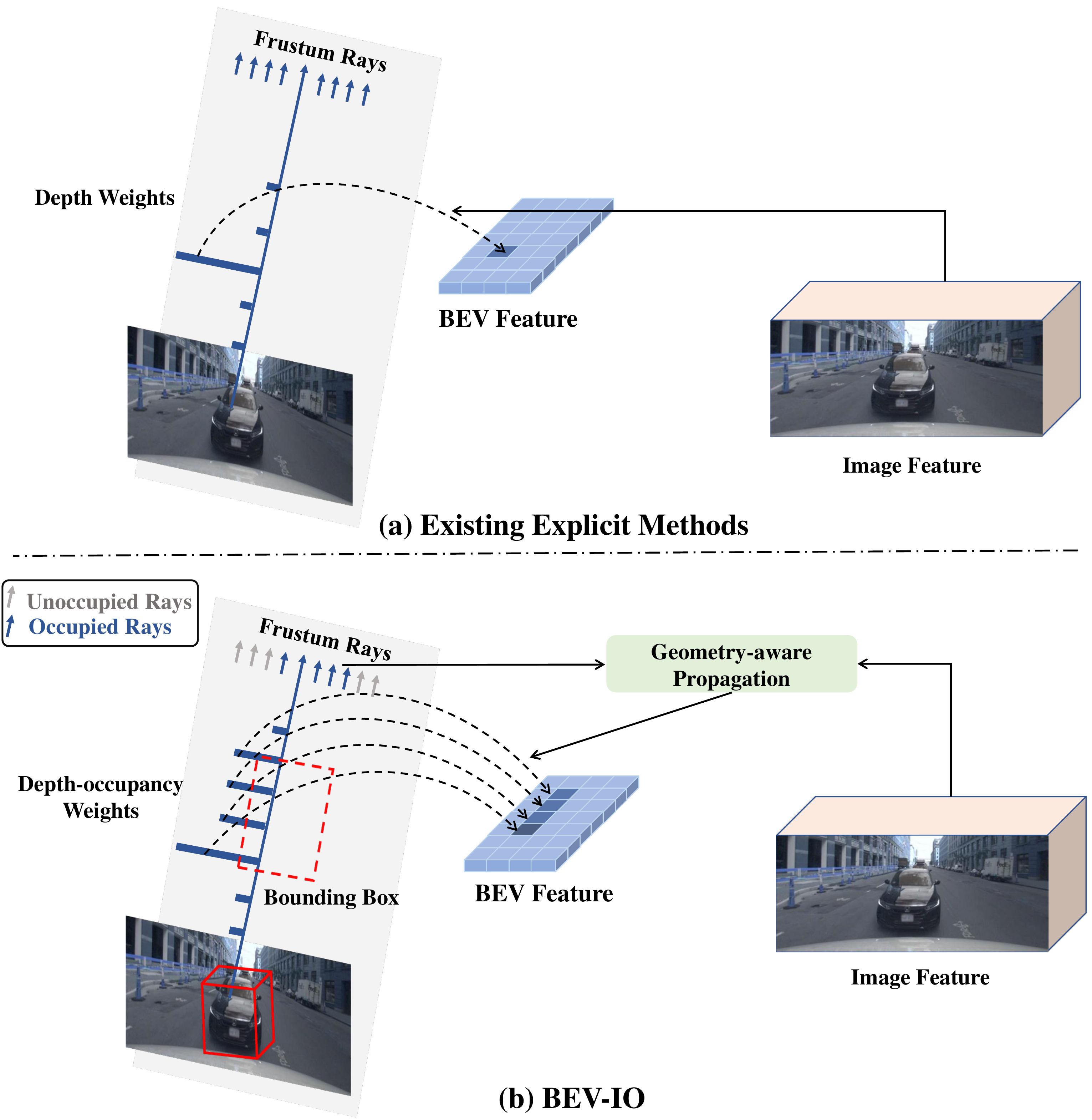}
    \end{center}
\vspace{-5mm}
    \caption{\textbf{Comparison of our BEV-IO with existing explicit BEV-based detection methods.} (a) Existing BEV-based methods utilize estimated depth weights to lift image features onto the BEV space, where the depth weights only feature visible surfaces. (b) We introduce the occupancy weights upon the depth weights to obtain more complete and precise BEV representations. In addition, we propose the propagation of image features with occupancy cues to attain geometry-aware image features.
    }
\label{fig:teaser}
\end{figure}

Recently, the adoption of bird's-eye-view (BEV) representation has become a prevalent solution for most 3D detection methods, enabling multi-view and cross-modality feature fusion in a unified and end-to-end manner. Although LiDAR-based~\cite{lang2019pointpillars, zhou2018voxelnet} and fusion-based~\cite{liu2022bevfusion, beidaliang2022bevfusion, Qin_2023_ICCV, li2022voxelfusion, jiao2023msmdfusion, yoo20203dcvf, li2022deepfusion, bai2022transfusion, chen2022autoalign} methods are capable of achieving reliable 3D detection, the associated costs of data acquisition can be exponentially high. 
In comparison, camera-based methods~\cite{bevdet, bevdepth, li2022bevstereo, bevformer, bevformerv2, Park2022solo, li2023fbbev, zhou2023matrixvt, wang2023distillbev, wang2023frustumformer} offer a more cost-effective alternative to LiDAR-based approaches, showcasing immense potential across various real-world applications like autonomous driving and robotics.
Camera-based methods can be categorized into two main streams that differ in whether they explicitly estimate scene depth distributions. Implicit methods~\cite{bevformer, bevformerv2, yang2023parametric, wang2023frustumformer} circumvent the need of depth estimation by employing BEV queries with cross-attention mechanisms, thereby implicitly generating the BEV features from multi-view images. Nevertheless, the absence of depth information in camera-based methods makes them susceptible to overfitting when confronted with corner cases.
In comparison, explicit methods~\cite{ECCV2020LSS, bevdepth, bevdet, li2022bevstereo} accomplish BEV representations by initially projecting 2D image features into the frustum space using estimated depth, followed by splatting these features onto the BEV space.

In this paper, we focus on explicit methods, where BEV feature construction is the essential step and involves two fundamental aspects, \emph{ie.}, ``\textbf{how to lift the 2D feature onto BEV space}" and ``\textbf{what features to lift}". 
The key to the first question is highly reliant on accurate 3D geometry perception. LSS~\cite{ECCV2020LSS} pioneered the 2D-BEV lifting process, implicitly learning depth via BEV bounding box ground truth. Due to the lack of depth ground truth, it mainly focuses on the depth of target areas, while exhibiting poor depth performance for the entire scene. Recent studies~\cite{bevdepth, li2022bevstereo, wang2022sts, Park2022solo, li2023fbbev} attempt to utilize sparse depth ground truth to supervise depth estimation, which significantly enhances the precision of the estimated depth and detection results. However, depth information only characterizes visible object surfaces and fails to capture their internal space or overall geometric structure, leading to sparse and unsatisfactory 3D representations in BEV space (See Fig.~\ref{fig:teaser} (a)). 

To address the above issues, we present BEV-IO, a new paradigm for 3D detection using instance occupancy information.
Occupancy indicates the probability of 3D points being occupied by objects, which, unlike depth, captures the comprehensive 3D spatial geometry information for the scene. Therefore, our core idea is to leverage instance occupancy prediction to assist explicit BEV-based detection methods through more complete and accurate feature lifting onto the BEV space (See Fig.~\ref{fig:teaser} (b)). To this end, we propose the instance occupancy prediction (IOP) module, which explores both explicit and implicit training manners. The explicit training manner mines strong supervisory signals from 3D bounding box annotations, while the implicit training strategy optimizes occupancy prediction with respect to the final detection performance in an end-to-end manner, maintaining more flexible training. By integrating the above two strategies, our IOP module is able to fill in object inner structures and is less affected by sparse depth ground truth, giving rise to 
more instance-oriented and comprehensive BEV representation.

To investigate the second question of \textbf{"what feature to lift"}, most existing methods solely utilize 2D image features, while ignoring the interaction between the 3D geometry and the 2D image domains. Recent works~\cite{liu2022bevfusion, beidaliang2022bevfusion} suggest that the fusion of geometry and image features is crucial for BEV-based detection. 
Rather than using complex 3D encoders to extract geometry features, we design an efficient geometry-aware feature propagation (GFP) mechanism as an alternative solution, which leverages instance geometry cues (occupancy) to propagate image context features. This is achieved through self-attention operations featured by occupancy distribution along each input ray to propagate the geometry cues to image features. As such, our GFP mechanism offers two unique advantages. For one thing, it can effectively enforce feature consistency within each object instance, as feature points within the same instance share similar occupancy statuses. For another, the feature propagation process also encodes the 3D information of target objects (\emph{e.g.}, orientation angles, bounding box size \emph{etc.}), leading to more accurate 3D detection results. 


By combining the IOP modules with the GFP mechanism, our BEV-IO method is compatible with most existing BEV-based frameworks and therefore has the potential to benefit prior 3D detection methods. Our main contribution can be summarized as follows:

\begin{itemize}
    \item We investigate the limitations of the depth-oriented feature lifting process in existing methods and propose a new instance occupancy prediction~(IOP) module to enhance the integrity and accuracy of the BEV feature representation.
    \item We propose an efficient geometry-aware feature propagation mechanism~(GFP) to further enhance image features with instance occupancy cues.
    \item Our BEV-IO achieves significant improvement under a popular BEV detection framework, while only adding $0.2\%$ parameters and $0.24\%$ GFLOPS.
\end{itemize}
Source code and models will be released to facilitate reproduction.


\section{Related works}

\subsection{Multi-view 3D Object Detection}
The objective of multi-view object detection is to detect 3D bounding boxes of objects in multi views. Existing methods follow a unified paradigm of projecting image features onto the BEV space for detection. These methods can be categorized into explicit~\cite{ECCV2020LSS,bevdepth,bevdet,bevdet4d} and implicit~\cite{bevformer, bevformerv2} approaches based on their different projection methods, through feature lifting and feature querying, respectively.

\textbf{Implicit BEV-based Detection}
Implicit methods do not rely on explicit geometric information but instead use attention mechanisms to obtain BEV features. BEVFormer~\cite{bevformer} utilizes pillars to perform deformable cross-attention between BEV queries and image features to obtain BEV features. Building on this, BEVFormerv2~\cite{bevformerv2} introduces a perspective detector to better transfer 2D object detection capabilities to 3D detection. PETR~\cite{PETR} proposes to enhance the geometry awareness of BEV features by incorporating positional encoding. Based on PETR, DA-BEV~\cite{dabev} proposes to further enhance the relationship between BEV spatial information and predicted depth. OA-BEV~\cite{oa-bev} proposes to utilize instance point cloud feature to enhance detection performance.
FrustumFormer~\cite{wang2023frustumformer} proposes an instance frustum which utilizes instance mask and BEV occupancy mask to enhance the feature interaction of instances. Different from the BEV occupancy mask in FrustumFormer, our approach directly predicts point-level occupancy in 3D space, which can better assist explicit methods in feature lifting process.

\begin{figure*}[!t]
    \centering
    \includegraphics[width=1\textwidth]{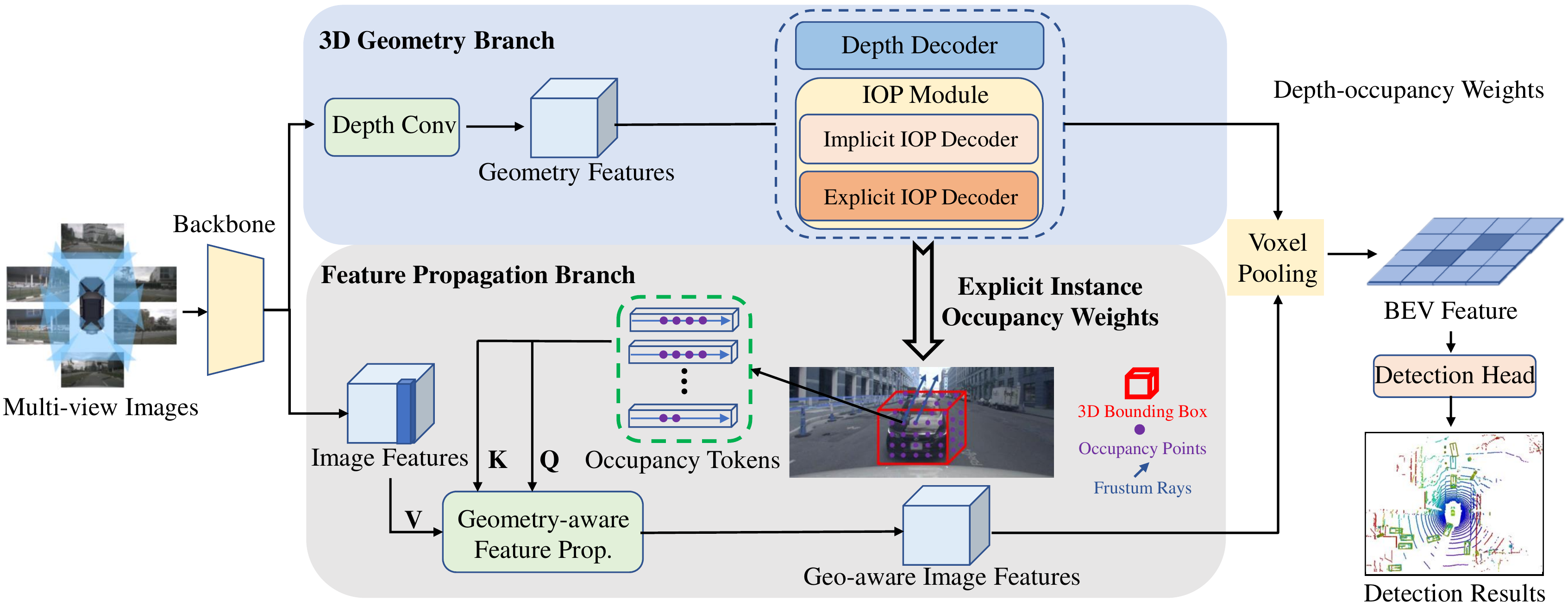}
    \caption{\textbf{Overall pipeline of our BEV-IO}. BEV-IO is mainly composed of a 3D geometry branch and a feature propagation branch. (1) 3D geometry branch utilizes the image features as input to estimate depth and explicit/implicit instance occupancy weights. Subsequently, these weights are fused to generate depth-occupancy weights. (2) The feature propagation branch takes image features and explicit instance occupancy weights as input, and a geometry-aware propagation module is performed to further enhance the image features with geometry cues. Finally, obtained geometry-aware features are lifted onto the BEV space using the depth-occupancy weights. Obtained BEV features are fed into the detection head to obtain the final detection results.
    }
\label{fig:pipeline}
\end{figure*}

\textbf{Explicit BEV-based Detection}
The core idea of explicit methods is to lift 2D image features onto the BEV space using geometric information. LSS~\cite{ECCV2020LSS} is the first explicit method that proposes to lift features onto the frustum space using depth weights and then splat them onto BEV space. BEVDet~\cite{bevdet} proposes a strong data augmentation strategy. Based on BEVDet, BEVDet4D~\cite{bevdet4d} introduces temporal information to achieve significant performance improvements. However, these methods do not utilize ground-truth depth supervision, thus lacking the geometry structure information of the entire scene. BEVDepth~\cite{bevdepth} uses ground-truth depth to improve depth estimation performance, significantly improving detection capabilities. Building on this, BEVStereo~\cite{li2022bevstereo} and STS~\cite{wang2022sts} further improve depth accuracy through the multi-view mechanism, leading to further performance gains. SOLOFusion leverages both long-term and short-term temporal fusion techniques to enhance performance. TIG-BEV~\cite{tigbev} proposes to enhance the representation ability of BEV features by estimating the relative depth within instances, which strengthens the perception of geometric information. The aforementioned approaches are all based on depth, however, the depth-based feature filling is not sufficiently comprehensive. 
Instead of depth, Simple-BEV~\cite{harley2023simplebev} proposes a parameter-free sampling method to comprehensively lift features onto BEV plane.
Different from the above depth and sampling methods, 
We propose utilizing depth-occupancy weights to aid in the accurate and comprehensive feature representation in the BEV space. 

\subsection{Occupancy Prediction}
Occupancy represents the occupancy status of spatial elements. MonoScene~\cite{cao2022monoscene} proposes to estimate depth frustum for occupancy prediction. OccDepth~\cite{miao2023occdepth} proposes estimating metric depth with stereo prior to enhance occupancy prediction. Voxformer~\cite{li2023voxformer} proposes a two-stage pipeline decoupling occupancy prediction and category prediction. TPVFormer~\cite{tpvformer} proposes utilizing triple-view features to estimate occupancy. OpenOccupancy~\cite{wang2023openoccupancy} and OpenOcc~\cite{openocc} propose dense annotation of occupancy on the nuscenes dataset. OccupancyM3D~\cite{peng2023learning3dwithocc} and Occ-BEV~\cite{min2023occbev} incorporate occupancy into BEV (Bird's Eye View) detection. Unlike the aforementioned works, our approach generates occupancy using the priors of bounding boxes, eliminating the need for additional occupancy annotation and focusing more on the collective information of the bounding boxes.

\section{Method}

\subsection{Explicit BEV Detection Revisit}

Camera-based multi-view BEV detection methods aim to detect object bounding boxes from the BEV perspective. First, the images from six views $\{I_i\in{R^{3\times{H}\times{W}}},i=1,2,...,6\}$ are fed into a pre-trained image encoder to obtain the image features. $\{f_i\in{R^{C\times{H}\times{W}}},i=1,2,...,6\}$, where $H$, $W$ and $C$ represent the height, width, and dimension of the context feature map, respectively. Different from the implicit methods, the explicit methods estimate the depth frustum for all input views $\{D_i\in{R^{C^D\times{H}\times{W}}},i=1,2,...,6\}$, where the $D_i$ represents probabilities for different handcrafted depth bins along each ray in the frustum and $C^D$ denotes the number of depth bins. The image feature frustum $F_i$ is achieved by weighting the image feature $f_i$ using the depth frustum $D_i$ , which can be mathematically described as the outer product between $f_i$ and $D_i$:
\begin{equation}
F_i=f_i\otimes{D_i},F_i\in{R^{C^{D}\times{C}\times{H}\times{W}}}.
\end{equation}
The image feature frustums of multiple views are then lifted onto the 3D space and splatted onto the BEV space through voxel pooling operation~\cite{ECCV2020LSS}:
\begin{equation}
    \begin{aligned}
        &F^{BEV}=\text{VoxPooling}(\{\text{Proj}\langle\ {F_i,K} \rangle\ ,i=1,2,...,6\}), \\
        &F^{BEV}\in{R^{{C}\times{H_v}\times{W_v}}},
    \end{aligned}
\end{equation}

Where $K$ denotes camera intrinsic, Proj is the projecting operation from image to 3D space, and VoxPooling is the voxel pooling process. 
The obtained BEV features are finally fed into the task-specific detection heads to obtain the final results of object class, locations, size and velocity.

\subsection{Overview of BEV-IO}
To maintain universality and compatibility with most prior methods, we instantiate our BEV-IO method by following the above prevalent BEV-based detection pipeline, which consists of three components: an image encoder, a view-transformer, and a detection head. We adopt the same image encoder and detection head of the popular BEVDepth detector~\cite{bevdepth}. As shown in Fig.~\ref{fig:pipeline}, our view-transformer comprises two branches, \emph{ie.}, a 3D geometry branch, and a feature propagation branch. (1) The 3D geometry branch consists of a depth decoder and two instance occupancy prediction (IOP) decoders. Given the image features from the image encoder, the depth decoder predicts depth weights, following the same structure as BEVDepth~\cite{bevdepth}, while the IOP decoders predict instance occupancy weights in the frustum space. Both depth and explicit IOP decoder are supervised by ground truth. The depth and occupancy weights are fused into the depth-occupancy weights and used to lift the image features onto the frustum space. (2) The core of the feature propagation branch is our geometry-aware feature propagation (GFP) module. It takes the explicit occupancy weights and image features as input and produces geometry-aware features, which will be lifted onto the frustum to obtain the BEV feature. In the following, we will delineate the above two branches.

\subsection{3D Geometry Branch}
\begin{figure*}[t]
    \centering
    \includegraphics[width=0.9\textwidth]{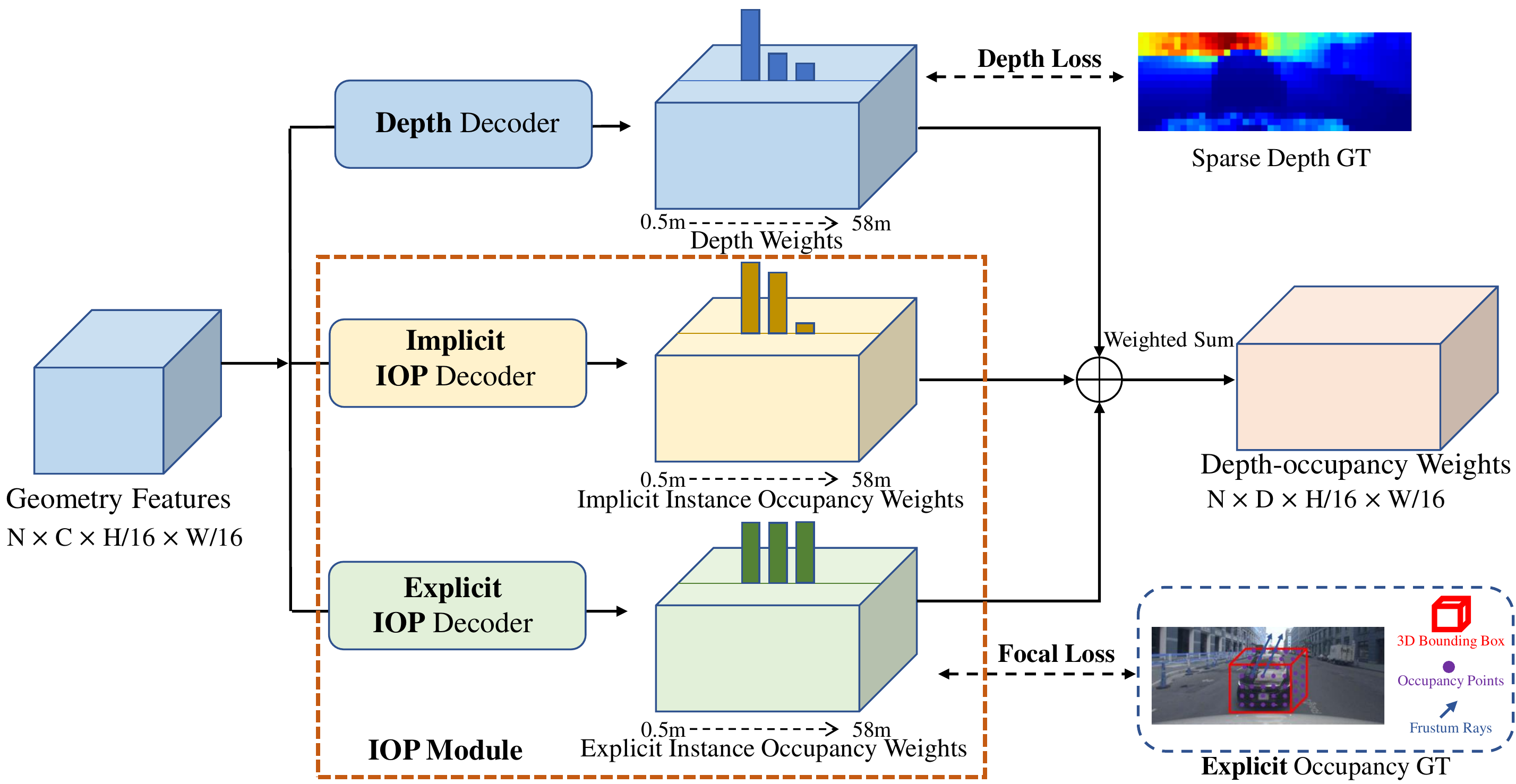}
    \caption{\textbf{Illustration of our 3D geometry branch}. The 3D geometry branch takes image features as input to predict depth, explicit and implicit instance occupancy weights. Depth and explicit instance occupancy are supervised by ground truth depth and generated ground truth explicit occupancy respectively. The implicit occupancy weights are only supervised by final detection loss. The depth-occupancy weights are the weighted sum of these three weights.
    }
\label{fig:3dgeo}
\end{figure*}

Existing explicit BEV detection methods~\cite{bevdepth, li2022bevstereo} lift the features onto the BEV space based on the estimated depth. However, depth information alone is not sufficiently comprehensive for the feature lifting process, as discussed in Sec.~\ref{sec:intro}. To address these issues, we propose to leverage point-level instance occupancy for feature lifting. Therefore, as shown in Fig.~\ref{fig:3dgeo}, our 3D geometry branch contains not only a depth decoder as in prior methods but also two IOP decoders in parallel to infer a more comprehensive 3D geometry of the scene.

\paragraph{Depth Decoder.} Following BEVdepth~\cite{bevdepth}, the depth decoder predicts depth weights $\{D_i\in{R^{C^D\times{H}\times{W}}},i=1,2,...,6\}$ for a set of manually designed depth bins, which is supervised using the following binary cross entropy loss:
\begin{equation}
    L_{\text{depth}} = \sum_{i=1}^{6} \text{BCE}(D_i, D^{gt}_i),
\end{equation}
where the depth ground truth $D^{gt}_i$ is transformed into the one-hot format during loss computation.

\paragraph{Implicit IOP Decoder.}
The depth weight only captures the visual surfaces of objects. To fill in the missing inner spaces of objects, we design the implicit IOP decoder, which estimates the occupancy probabilities of the depth bins along each ray $\{O_i^{im}\in{R^{C^D\times{H}\times{W}}},i=1,2,...,6\}$. As opposed to the depth decoder supervised by ground truth depth, we train the implicit IOP decoder without explicit supervision but rather in an end-to-end manner towards the ultimate goal of 3D detection. As such, the implicit IOP decoder enjoys strong flexibility and has the potential to generate the optimal spatial weights for image feature lifting that can finally yield more accurate 3D detection results.    

\paragraph{Explicit IOP Decoder.}
The implicit IOP decoder without direct and concrete supervision may be hard to train. To remedy this issue, we further design an explicit IOP decoder to predict explicit occupancy weights $\{O_i^{ex}\in{R^{C^D\times{H}\times{W}}},i=1,2,...,6\}$. Considering that point-level occupancy is expensive to manually annotate and is very rare in existing datasets~\cite{wang2023openoccupancy}, we develop an effective and efficient approach to construct occupancy labels from 3D bounding box ground truth, which will be detailed later. For simplicity, we formulate occupancy labeling as a binary classification problem in the view-frustum space, where 3D points within an object bounding box will be labeled 1, and 0 otherwise. Since we aim to detect 3D bounding boxes, the above occupancy labeling strategy is therefore reasonable and consistent with our ultimate goal while eliminating the need for expensive manual annotation. During training, we adopt the following focal loss~\cite{lin2017focal} to mitigate the effects of class imbalance:
\begin{equation}
    L_{\text{exocc}} = \sum_{i=1}^{6} \text{FocalLoss}(O^{ex}_i, O^{gt}_i),
\end{equation}
where $O^{gt}$ denotes the binary occupancy labels. Given the predicted implicit and explicit occupancy weights as well as the depth weights, we obtain the final depth-occupancy weights as follows:
\begin{equation}
S_i = w^d D_i + w^{im} O^{im}_i + w^{ex} O^{ex}_i,
\end{equation}
where $w^d$, $w^{im}$, and $w^{ex}$ are trainable parameters for weighted combination. The final depth-occupancy weights are used to lift 2D image features onto the view-frustum space.

\paragraph{Point-level Occupancy Ground Truth Generation.}
We present a simple but effective method to label point-level occupancy based on 3D object bounding boxes.
Specifically, for each bounding box, we partition the 3D space into two parts: the interior and the exterior of the box. For each point, we determine its position relative to the box by calculating the dot product between the point and the surface normal vectors of the six box faces. If the point falls within all six faces, it is deemed occupied. The algorithmic process is summarized in Algorithm~\ref{alg}.

\begin{algorithm}[t]
    \SetKwInOut{Input}{Input}\SetKwInOut{Output}{Output}
    
    \Input{Frustum points set $P$; 3D bounding box $B$}
    \Output{Occupied points set $P_o$}
    \caption{Point-level occupancy ground truth generation}\label{alg}
    
    Initialize $P_o=\emptyset$\;
    \For{each point $p$ in points set $P$}{
    Initialize $s=0$\;
      \For{each face $F$ of 3D bounding box $B$}{
        Compute surface normal vector $\vec{n}$ of face $F$\;
        Select a point $q$ on the face $F$\;
        Compute vector $\vec{v} = \vec{pq}$\;
        Compute dot product $d = \vec{v} \cdot \vec{n}$\;
        \uIf{$d < 0$}{
        $s\leftarrow s+1$
        }
        }
    \uIf{$s == 6$}{
    Add point $p$ into occupied points set $P_o$
    }
      }
\end{algorithm}

\begin{figure*}[t]
    \centering
    \includegraphics[width=0.9\textwidth]{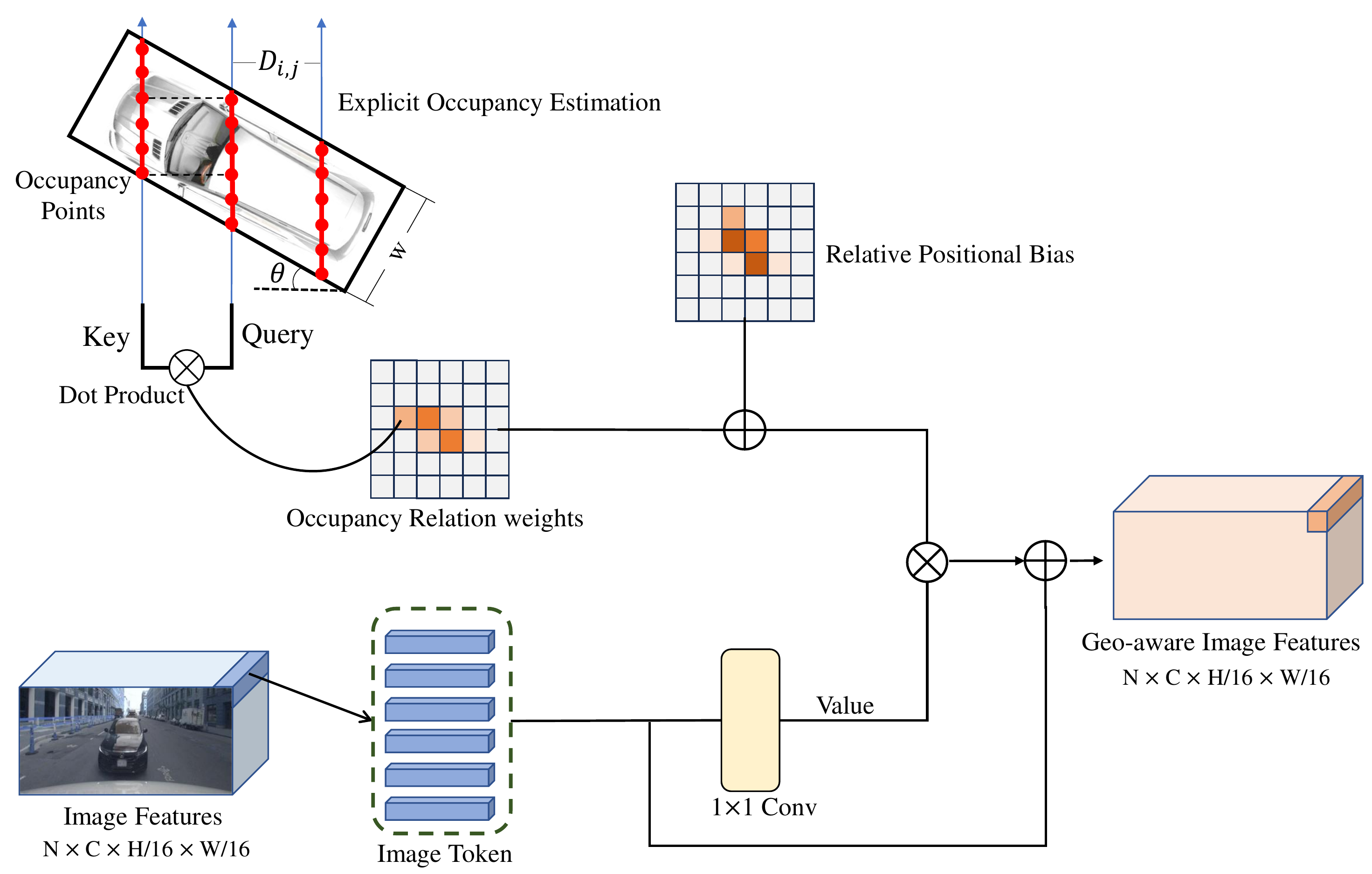}
    \caption{\textbf{Geometry-aware feature propagation mechanism.} The input of geometry-aware feature propagation mechanism (GFP) is the predicted explicit instance occupancy and image features. GFP takes the explicit occupancy tokens as the key and value, performs self-attention, and conducts geometry-aware feature propagation on the image features.
    }
\label{fig:occmodule}
\end{figure*}

\subsection{Geometry-aware Feature Propagation Branch}

The predicted instance occupancy weights along each ray encode the geometric structure information of the scene. Moreover, feature points on the same object share similar occupancy weights. Motivated by this insight, we 
design a geometry-aware feature propagation~(GFP) module to enforce feature consistency within each object region by leveraging the geometric information of occupancy weights. As shown in Fig.~\ref{fig:occmodule}, we use the explicit occupancy weights of each feature point as the key and query, and its image feature as values. We then calculate the self-attention to achieve geometry-aware propagation of image features.  

Through a more in-depth analysis (See supplementary materials for details), the attention weights $a_{i,j}$ between two feature points located on the same target instances can be geometrically calculated as follows:
\begin{equation}
    a_{i, j} = O_i^{ex} \cdot O_j^{ex} = \frac{w}{\cos{\theta}}-D_{i, j}{\tan{\theta}}, 
\end{equation}
where $w$ is the width of the target bounding box, $D_{i, j}$ is the spatial distance between the two feature points, and ${\theta}$ denotes the bounding box orientation angle as shown in Fig.~\ref{fig:occmodule}. It can be observed that the attention weights encode the geometry information (3D size and orientation angles) of the target.
As a consequence, the generated image features of GFP become not only spatially more consistent but also geometry-aware. As demonstrated in the experiments (Tab.~\ref{tab:ablation_study}), the proposed GFP can substantially improve the final detection accuracy.

\subsection{Loss Function}
The final loss function is a weighted combination of the detection loss $L_{\text{det}}$, depth loss $L_{\text{depth}}$, and occupancy loss $L_{\text{exocc}}$ as follows:
\begin{equation}
    L_{\text{total}} = {\lambda}_{1}{L_{\text{det}}} + {\lambda}_{2}{L_{\text{depth}}} + {\lambda}_{3}{L_{\text{exocc}}},
\end{equation}
where we adopt the loss function in \cite{bevdet} as the detection loss $L_{\text{det}}$, and hyper-parameters ${\lambda}_{1}$, ${\lambda}_{2}$, and ${\lambda}_{3}$ are pre-defined to balance the three loss terms.

\section{Experiment}

\begin{table*}[t]
  \centering
  
  \resizebox{1.0\linewidth}{!}{
    \begin{tabular}{l|cc|cc|ccccc}
    \hline

    \hline
    Methods                     &Image Size      & Backbone & \textbf{mAP}$\uparrow$ & \textbf{NDS}$\uparrow$ & mATE$\downarrow$ & mASE$\downarrow$   & mAOE$\downarrow$  & mAVE$\downarrow$  &  mAAE$\downarrow$   \\
    \hline
    BEVDet \cite{bevdet} &704$\times$256 & ResNet50 & 0.298   & 0.379 &0.725             & 0.279            & 0.589             & 0.860             & 0.245              \\
    PETR \cite{PETR}& 1056$\times$384                         & ResNet50    &0.313          &0.381 &0.768              &0.278             &0.564             &0.923             &0.225             \\
    BEVDet4D \cite{bevdet4d}& 704$\times$256                        & ResNet50    &0.322          &0.457 &0.703             &0.278               &0.495             &0.354             &0.206             \\
    BEVDepth \cite{bevdet4d}& 704$\times$256                        & ResNet50    &0.351          &0.475 &0.639             &0.267               &0.479             &0.428             &0.198             \\
    STS \cite{wang2022sts}& 704$\times$256                        & ResNet50    &0.377          &0.489 &0.601             &0.275               &0.450             &0.446             &0.212             \\
    BEVStereo \cite{li2022bevstereo}& 704$\times$256                        & ResNet50    &0.372          &0.500 &0.598             &0.270               &0.438             &0.367             &0.190             \\
    SOLOFusion \cite{Park2022solo}& 704$\times$256                        & ResNet50    &0.427         &0.534 &0.567             &0.274               &0.411             &0.252             &0.188             \\
    \hline
    BEVDet4D-Depth$\dag$\cite{bevdet4d}     &704$\times$256          & ResNet50   & 0.355         & 0.480     & 0.641        & 0.280              & 0.464             & 0.385             & \textbf{0.203}               \\
    \textbf{Ours}               &704$\times$256             & ResNet50   & \textbf{0.368}         & \textbf{0.493}      & \textbf{0.611}       & \textbf{0.271}              & \textbf{0.463}             & \textbf{0.359}             & 0.207                \\
                   &             &    & $\color{blue}\textbf{+1.3\%}$        & $\color{blue}\textbf{+1.3\%}$      & $\color{blue}\textbf{-3.0\%}$       & $\color{blue}\textbf{-0.9\%}$              & $\color{blue}\textbf{-0.1\%}$            & $\color{blue}\textbf{-2.6\%}$            & $\color{red}\textbf{+0.4\%}$                \\
    
    \hline
    BEVDet4D-LongTerm$\dag$\cite{bevdet4d}     &704$\times$256          & ResNet50   & 0.387        & 0.511     & 0.579      & 0.278              & 0.477             & 0.295             & 0.193              \\
    \textbf{Ours}               &704$\times$256             & ResNet50   & \textbf{0.401}         & \textbf{0.524}      & \textbf{0.563}       & \textbf{0.272}              & \textbf{0.454}             & \textbf{0.287}             & \textbf{0.192}                \\
    &             &    & $\color{blue}\textbf{+1.5\%}$        & $\color{blue}\textbf{+1.3\%}$      & $\color{blue}\textbf{-1.6\%}$       & $\color{blue}\textbf{-0.6\%}$              & $\color{blue}\textbf{-2.3\%}$            & $\color{blue}\textbf{-0.8\%}$            & $\color{blue}\textbf{-0.1\%}$                \\
    \hline
    DETR3D\cite{DETR3D}  &  1600$\times$900                            & ResNet101   & 0.349         & 0.434 & 0.716            & 0.268              & 0.379             & 0.842             & 0.200               \\
    PETR\cite{PETR}  &  1408$\times$512                            & ResNet101   & 0.357         & 0.421 & 0.710            & 0.270              & 0.490             & 0.885             & 0.224               \\
    BEVDepth\cite{bevdepth}  &  1408$\times$512                            & ResNet101   & 0.412         & 0.535 & 0.565            & 0.266              & 0.358             & 0.331             & 0.190               \\
    STS \cite{wang2022sts}  &  1408$\times$512                            & ResNet101   & 0.431         & 0.542 & 0.525            & 0.262              & 0.380             & 0.369             & 0.204               \\
    SOLOFusion \cite{Park2022solo}  &  1408$\times$512                            & ResNet101   & 0.483         & 0.582 & 0.503            & 0.264              & 0.381             & 0.246             & 0.207               \\

    \hline
    
    BEVDet4D-Depth$\dag$\cite{bevdet4d}     &1408$\times$512          & ResNet101   & 0.417     & 0.518           & 0.622            & 0.271             & 0.423            & 0.369             & 0.221  \\
    \textbf{Ours}               &1408$\times$512             & ResNet101   & \textbf{0.423}         & \textbf{0.534}             & \textbf{0.590}              & \textbf{0.272}             & \textbf{0.382}            & \textbf{0.332}             & \textbf{0.201}   \\
    &             &    & $\color{blue}\textbf{+0.5\%}$        & $\color{blue}\textbf{+1.6\%}$      & $\color{blue}\textbf{-3.2\%}$       & $\color{blue}\textbf{-0.1\%}$              & $\color{blue}\textbf{-4.1\%}$            & $\color{blue}\textbf{-3.7\%}$            & $\color{blue}\textbf{-2.0\%}$                \\

    \hline
    \end{tabular}%
    }
\caption{\textbf{Comparison of other state-of-the-art methods on the nuScenes \texttt{val} set}. "$\dag$" indicates our re-implementation of BEVDet codebase~\cite{bevdet}. BEVDet4D-Depth indicates using the same method with BEVDepth~\cite{bevdepth} without the depth aggregation module. BEVDet4D-LongTerm indicates using the long-term fusion method proposed in SOLOFusion~\cite{Park2022solo}.
      }    
\label{tab:nus-val}
\end{table*}%

\subsection{Datasets and Metrics}
\textbf{Nuscenes Dataset.} We evaluate BEV-IO on nuScenes dataset~\cite{caesar2020nuscenes},
which is a large-scale autonomous driving dataset containing over 1,000 scenes of complex urban driving scenarios with up to 1.4 million annotated 3D bounding boxes for 10 classes.
The scenes are recorded in Boston and Singapore, and include a wide range of weather and lighting conditions, as well as various traffic scenarios. All the sequences are officially split into 700/150/150 for training, validation, and test. 

\textbf{Evaluation Matrix.} We utilize the official evaluation criteria in NuScenes including nuScenes Detection Score (NDS) and mean Average Precision (mAP), along with mean Average Translation Error (mATE), mean Average Scale Error (mASE), mean Average Orientation Error (mAOE), mean Average Velocity Error (mAVE), and mean Average Attribute Error (mAAE).

\begin{table*}[t]
  \centering
  	\resizebox{0.9\linewidth}{!}{
    \begin{tabular}{c|ccc|cc|ccccc}
    \hline

    \hline
    ~ &Im-Occ                     &Ex-Occ      & GFP & \textbf{mAP}$\uparrow$ & \textbf{NDS}$\uparrow$ & mATE$\downarrow$ & mASE$\downarrow$   & mAOE$\downarrow$  & mAVE$\downarrow$  &  mAAE$\downarrow$  \\
    \hline
     (a)& ~ & ~ & ~                             & 0.331   & 0.438 & 0.684 & 0.274  & 0.563  & 0.524  & 0.231   \\
    (b)&\checkmark & ~ & ~                     & 0.338   & 0.451 & 0.679 & 0.274  & 0.529  & 0.492  & \textbf{0.201}   \\
    (c)& ~ & \checkmark & ~                     & 0.340   & 0.454 & 0.680 & 0.275    & 0.527  & 0.453 & 0.223   \\
    (d) & \checkmark& \checkmark & ~             & 0.342   & 0.459 & 0.669 & 0.273  & 0.530  & \textbf{0.428}  & 0.221   \\
    (e) & \checkmark& \checkmark & \checkmark    & \textbf{0.344}   & \textbf{0.463} & \textbf{0.660} & \textbf{0.267}  & \textbf{0.510}  & 0.431  & 0.220    \\
    \hline
    \end{tabular}%
    }
  \caption{\textbf{Performance of the key components in BEV-IO.}}
  \label{tab:ablation_study}%
\end{table*}%

\begin{figure*}[t]
    \centering
    \includegraphics[width=0.9\textwidth]{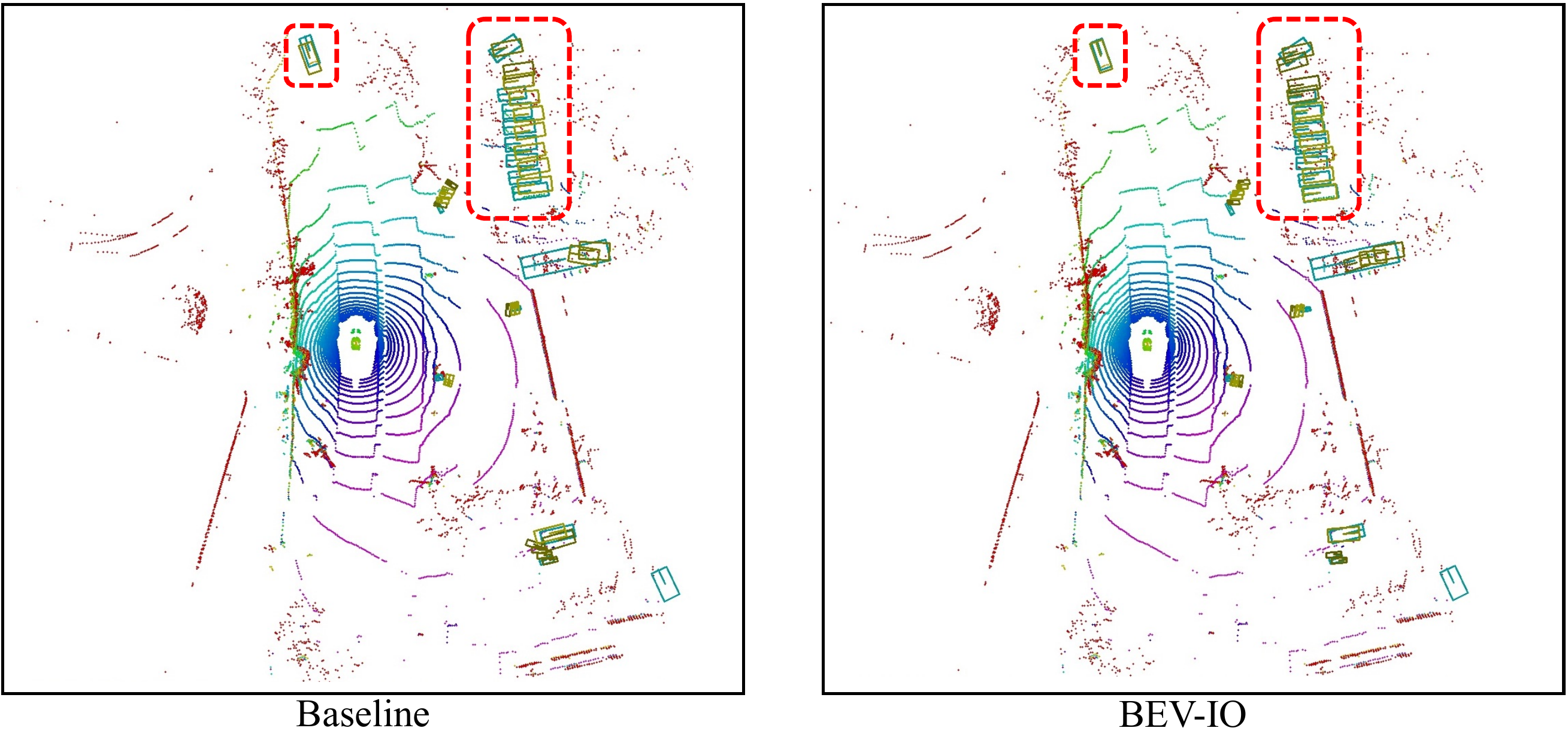}
    \caption{\textbf{Visualization of detection performance.} 
    The predicted and ground truth bounding boxes are marked in yellow and green, respectively. BEV-IO obtains more accurate predictions compared with the baseline method as shown in the red dashed boxes. 
    }
\label{fig:visual_detection}
\end{figure*}

\subsection{Experiment Settings}
Our experiments are conducted based on BEVDepth~\cite{bevdepth}. 
We adopt ResNet-50~\cite{resnet} as the image backbone and resize the image resolution to $704 \times 256$. Image and BEV data augmentation are adopted following BEVDet~\cite{bevdet,bevdepth}, including random cropping, random scaling, random flipping, and random rotation. 
During the training phase, 
only the ground truth of projected depth and detection annotations are utilized as supervision. AdamW~\cite{adamW} optimizer is used with a learning rate of 2e-4 and a batch size of 32. The loss weights $\lambda_{1}$, $\lambda_{2}$ and $\lambda_{3}$ are $1.0$, $3.0$ and $3000.0$ respectively. 
The CBGS~\cite{CBGS} strategy is applied during network training for a fair comparison with other methods, which is disabled in our ablation studies to improve training efficiency.
All experiments are conducted on 8 NVIDIA-3090 GPUs. 
At inference, our model receives multi-view images and camera intrinsic matrices and outputs the results of object class, locations, size, and velocity.

\subsection{3D Object Detection Results}
We compared our method with other BEV-based methods on the NuScenes val set, all of which are trained using the CBGS~\cite{CBGS} strategy. As shown in Tab.~\ref{tab:nus-val}, our method outperforms BEVDepth with a significant margin while only adding 0.15M parameters and 0.6 GFLOPS (shown in Tab.~\ref{tab:ablation_study_para}). The results with high-resolution input images are reported in the supplementary material.

\subsection{Ablation Study}
In this section, we conduct a series of ablation experiments to validate the effectiveness of the core components of our BEV-IO. The CBGS~\cite{CBGS} training strategy is disabled for all the compared models. Evaluation is performed using the nuScenes \textit{val} dataset.

\subsubsection{Ablation of BEV-IO}

Tab.~\ref{tab:ablation_study} reports the performance of the key components in BEV-IO. 
Im-Occ, Ex-Occ, and GFP refer to implicit instance occupancy prediction, explicit instance occupancy prediction, and geometry-aware feature propagation, respectively. The baseline setting, denoted as (a), can be regarded as BEVDepth~\cite{bevdepth}. In (b) and (c), Im-Occ and Ex-Occ are adopted, respectively, showing that both implicit and explicit instance occupancy prediction can improve performance compared with (a). The integration of Im-Occ and Ex-Occ in (d) can bring further benefit. Finally, (e) adds GFP upon (d) and achieves superior performance. These results suggest that all three components are crucial to the final detection performance.

\subsubsection{Ablation of Parameter Efficiency}
 
We further analyze the parameter and computational complexity comparison between our BEV-IO and the baseline method (\emph{i.e.}, BEVDepth). Tab.~\ref{tab:ablation_study_para} shows that BEV-IO can achieve significant improvement while adding only $0.2\%$ parameters and $0.24\%$ GFLOPs, which involves a very simple network comprising a few convolutional layers. Please see the supplementary material for the specific details of the network structure.
Fig.~\ref{fig:visual_detection} visualizes the detection results of the baseline and our BEV-IO. The predicted bounding boxes and the ground truth ones are denoted in yellow and green, respectively. Compared to the baseline method, BEV-IO obtains more accurate detection performance, whose predicted bounding boxes are better aligned with the ground truth (see the red dashed boxes).

\begin{table}[h]
  \centering
  	\resizebox{0.95\linewidth}{!}{
    \begin{tabular}{c|c|cc|cc}
    \hline

    \hline
    ~&Method & Param. & GFLOPs & \textbf{mAP}$\uparrow$ & \textbf{NDS}$\uparrow$  \\
    \hline
    (a)&Baseline & 74.9M & 252.4                            & 0.331   & 0.438  \\
    (b)&BEV-IO & 75.05M & 253.0          & \textbf{0.344}   & \textbf{0.463} \\
    \hline
    \end{tabular}%
    }
  \caption{\textbf{Comparsion with the baseline method on model complexity.}}
  \label{tab:ablation_study_para}%
\end{table}%

\section{Conclusion}
We propose BEV-IO, which leverages point-level instance occupancy to address the limitation of depth in capturing the entire instance. We design an instance occupancy prediction~(IOP) module that explicitly and implicitly estimates instance point-level occupancy to facilitate the more comprehensive BEV feature representation. Additionally, we introduce an geometry-aware feature propagation mechanism~(GFP) to effectively propagate image features by incorporating geometry cues. Experimental results demonstrate that our method outperforms state-of-the-art methods while only adding a negligible increase in parameters ($0.2\%$) and computational overhead ($0.24\%$ in GFLOPs).

\clearpage
{
    \small
    \bibliographystyle{ieeenat_fullname}
    \bibliography{main}
}


\end{document}


\maketitle

\section{Module Architecture}
~~Our depth-occupancy head and GFP network structure are illustrated in Fig~\ref{fig:network}. The additional network parameters of BEV-IO only include several 1x1 convolutions. Consequently, the resulting increase in parameter and computation overhead is minimal.

\begin{figure}[h]
    \centering
    \includegraphics[width=0.5\textwidth]{sup.pdf}
    \captionsetup{justification=centering}
    \caption{\textbf{Network architecture.}
    }
\label{fig:network}
\end{figure}

\section{In-Depth Analysis of GFP}
\begin{figure}[H]
    \centering
    \includegraphics[width=0.5\textwidth]{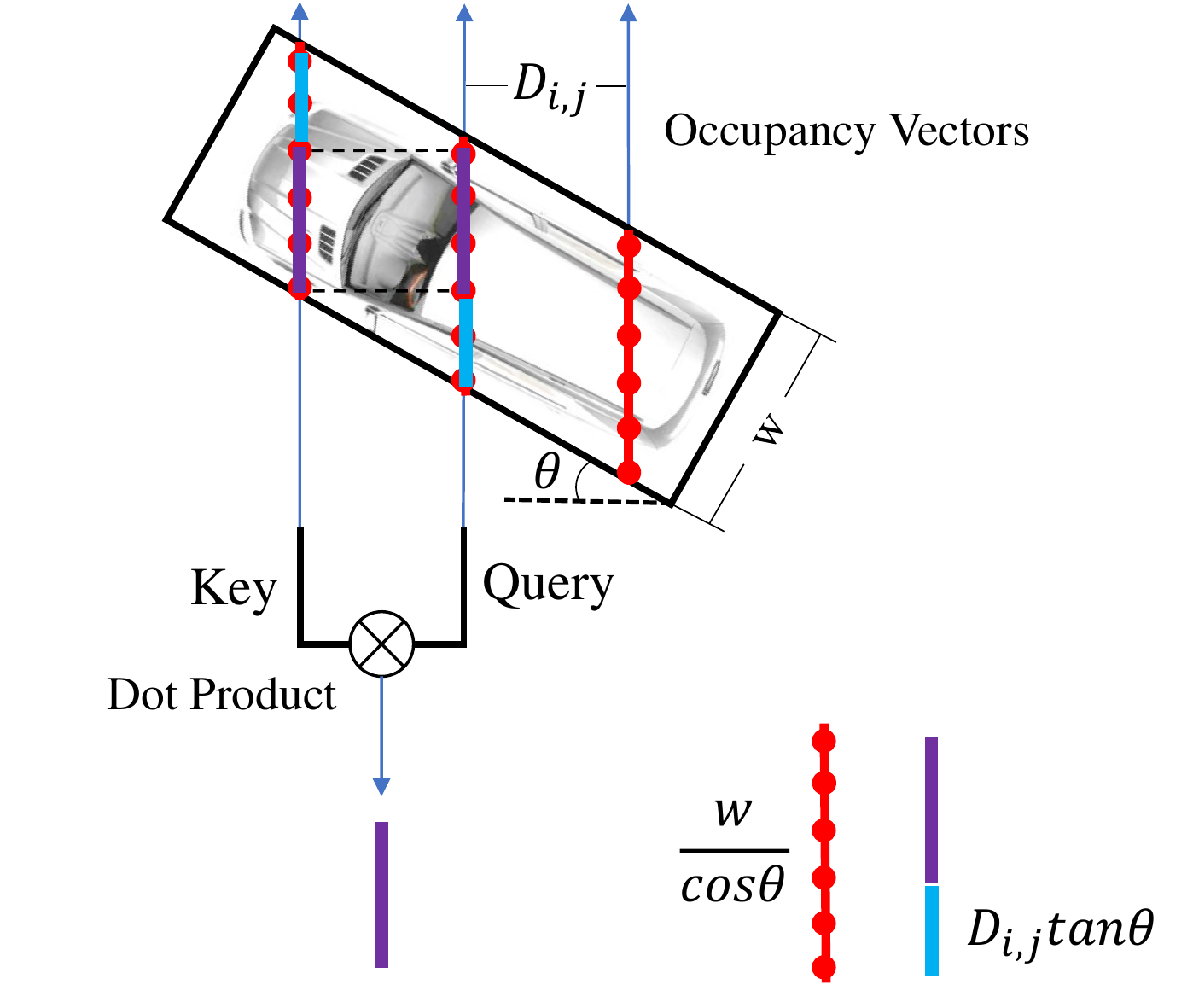}
    \captionsetup{justification=centering}
    \caption{\textbf{Dot product between occupancy vectors.}
    }
\label{fig:gfp}
\end{figure}

~~To enhance the utilization of spatial information, we employ the Geometry-Aware Feature Propagation module~(GFP). We utilize the dot product to characterize the relationships between different occupancy vectors and obtain attention values. As illustrated in Fig~\ref{fig:gfp}, the dot product of two occupancy vectors can be represented by the length of the overlapping area between them, which is the purple region in the image. The length of the purple region is equal to the length of the red region minus the length of the blue region.

